\documentclass{article}
\usepackage{ijcai24}

\usepackage{times}
\usepackage{soul}
\usepackage{url}
\usepackage[hidelinks]{hyperref}
\usepackage[utf8]{inputenc}
\usepackage[small]{caption}
\usepackage{graphicx}
\usepackage{amsmath}
\usepackage{amsthm}
\usepackage{booktabs}
\usepackage{algorithm}
\usepackage{algorithmic}
\usepackage[switch]{lineno}
\usepackage{algorithm}
\usepackage{algorithmic}
\usepackage{multirow} 
\usepackage{soul}
\usepackage{url}
\usepackage{anyfontsize}
\usepackage{booktabs}
\usepackage{algorithm}
\usepackage{algorithmic}
\usepackage[switch]{lineno}
\usepackage{amsfonts}
\usepackage{amssymb}
\usepackage{indentfirst}
\usepackage{bbding}
\usepackage{colortbl}
\usepackage{amsmath}
\usepackage[table]{xcolor}
\usepackage{graphicx} 
\usepackage{natbib}  
\usepackage{caption}

\title{Feature Denoising Diffusion Model for Blind Image Quality Assessment}

\author{
Xudong Li$^1$
\and
Jingyuan Zheng$^2$\and
Runze Hu$^{3}$\and
Yan Zhang\textsuperscript{\rm 1,}\footnote{Corresponding author}\and
Ke Li$^4$\and
Yunhang Shen$^4$\and \\
Xiawu Zheng$^1$\and
Yutao Liu$^5$\and
ShengChuan Zhang$^1$\and
Pingyang Dai$^1$\and
Rongrong Ji$^1$
\\
\affiliations
$^1$Key Laboratory of Multimedia Trusted Perception and Efficient Computing,\\ Ministry of Education of China, Xiamen University
$^2$School of Medicine, Xiamen University\\
$^3$School of Information and Electronics, Beijing Institute of Technology\\
$^4$Tencent Youtu Lab 
$^5$ School of Computer Science and Technology, Ocean University of China\\
\emails
\{lxd761050753, jyzheng0606,
bzhy986, hrzlpk2015, shenyunhang01, tristanli.sh\}@gmail.com,
liuyutao@ouc.edu.cn,\{zhengxiawu, zsc\_2016, pydai, rrji\}@xmu.edu.cn
}

\begin{document}

\maketitle

\begin{abstract}
Blind Image Quality Assessment (BIQA) aims to evaluate image quality in line with human perception, without reference benchmarks. Currently, deep learning BIQA methods typically depend on using features from high-level tasks for transfer learning. However, the inherent differences between BIQA and these high-level tasks inevitably introduce noise into the quality-aware features.
In this paper, we take an initial step towards exploring the diffusion model for feature denoising in BIQA, namely {Perceptual Feature Diffusion for IQA (PFD-IQA)}, which aims to remove noise from quality-aware features. Specifically, ({i})~We propose a {Perceptual Prior Discovery and Aggregation module} to establish two auxiliary tasks to discover potential low-level features in images that are used to aggregate perceptual text conditions for the diffusion model. ({ii}) We propose a {Perceptual Prior-based Feature Refinement strategy}, which matches noisy features to predefined denoising trajectories and then performs exact feature denoising based on text conditions. Extensive experiments on eight standard BIQA datasets demonstrate the superior performance to the state-of-the-art BIQA methods, i.e., achieving the PLCC values of 0.935 (\textcolor{red}{$\uparrow 3.0\%$} vs. 0.905 in KADID) and 0.922 (\textcolor{red}{$\uparrow 2.8\%$} vs. 0.894 in LIVEC).
\end{abstract}
\begin{figure}[htbp]
  \centering
    \includegraphics[width=0.47\textwidth]{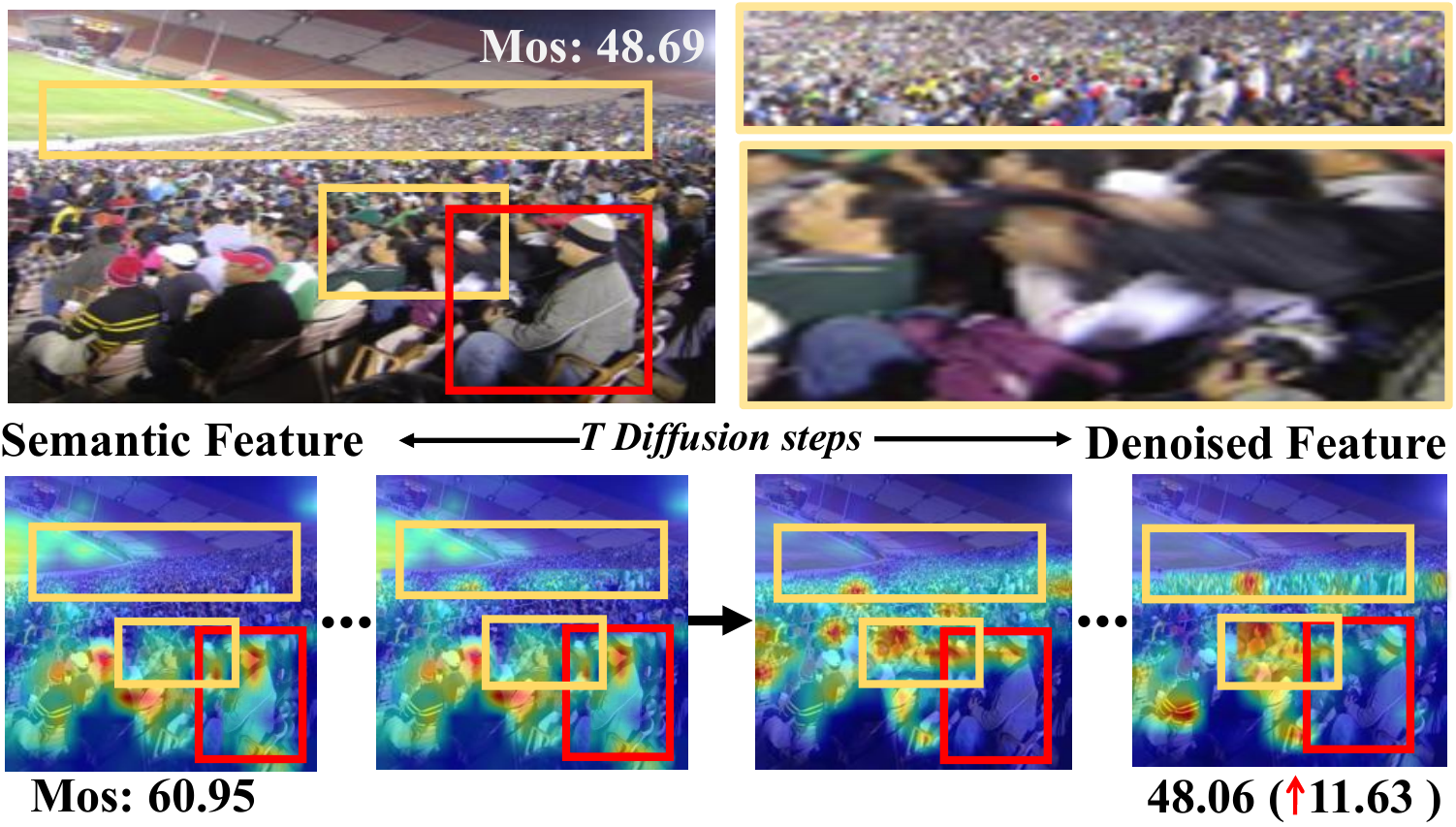}
  \caption{\textbf{Image on top:} the sample image. \textbf{Images at bottom:} Before and after diffusion denoising, the feature map significantly refines, effectively pinpointing areas with visible image quality degradation. The initial semantic focus is on ``human,`` but after denoising, attention notably shifts to the fuzzy region (the orange region with the blurred crowd and arms), resulting in a closer alignment with the actual Mean Opinion Scores (MOS). }
  \label{fig1}
\end{figure}
\section{Introduction}
Image Quality Assessment (IQA) methods aim to match the human perception of image distortions~\citep{wang2004image}. Reliable IQA models are important for image-driven applications, while also serving as benchmarks for image processing. Objective IQA includes Full-Reference IQA (FR-IQA)~\citep{shi2020full}, Reduced-Reference IQA (RR-IQA)~\citep{tao2009reduced}, and no-reference or Blind IQA (BIQA)~\citep{no}. As reference images are often unavailable, BIQA gains attention for tasks like image restoration~\citep{banham1997digital} and super-resolution~\citep{dong2015image} without references.

Data-driven BIQA models based on deep neural networks \citep{bosse2017deep, end} have made significant progress. The quality score of distorted images is typically measured using the Mean Opinion Score (MOS), making BIQA a small-sample task. To address this, a promising strategy utilizes a pre-training and fine-tuning paradigm, transferring shared features from the large-scale ImageNet source domain to the IQA target domain to accomplish the IQA task. However, during the pre-training of large-scale classification tasks, synthetic distortions are typically used as a data augmentation method, inevitably reducing the sensitivity of the model to image distortion~\citep{zhang2023blind,hendrycks2019benchmarking}. Consequently, the pre-trained features' insensitivity to distortion degradation can lead the model to excessively concentrate on high-level information during quality assessment, overlooking distortions information critical to quality perception~\citep{QPT,zhang2023blind}.
We provide an example to explain such a problem in Fig.~\ref{fig1}. As we observe, the baseline focuses excessively on the high-level information (such as semantic information, i.e., ``human`` in the red box) in the foreground of the distorted image, while neglecting the low-level quality-relevant information (such as the blur and geometric distortion in the yellow box), leading to inaccurate quality predictions. Therefore, these pre-trained features are not always beneficial, and some may even be considered as noise in the quality-aware features. It is necessary to meticulously filter out the noisy features.


The diffusion model \citep{ho2020denoising,rombach2022high} defines a Markov chain where noise is gradually added to the input samples (forward process) and then used to remove the corresponding noise from the noisy samples (reverse process), showcasing its effective noise removal ability. Inspired by this, a novel BIQA framework based on the diffusion model is first proposed, namely the Perceptual Feature Diffusion model for IQA (PFD-IQA). We formulate the feature noise filtering problem as a progressive feature denoising process, enabling effective enhancement of quality-aware features.
However, there are two challenges in directly utilizing diffusion models for denoising in BIQA:
(i) Traditional diffusion models may offer limited control over quality-aware feature preservation and noise feature elimination, possibly leading to suboptimal denoising.
(ii) In BIQA, explicit benchmarks or ground truths are often absent for denoising targets. This makes it challenging to define a clear denoising trajectory for the diffusion model.


To this end, our PFD-IQA consists of two main modules to overcome the above two challenges in the diffusion model for BIQA.
To address the problem (i), we introduce a Perceptual Prior Discovery and Aggregation module that merges text prompts representing various quality perceptions to guide the diffusion model. Specifically, we initially acquire potential distortion-aware and quality-level priors through auxiliary tasks. These are then combined based on their similarity to text prompts, creating perceptual text prompts, which serve as conditions to guide the model for more accurate feature denoising.
We select these priors for text descriptions primarily for two reasons: Firstly, understanding the diversity in distortion enhances prediction accuracy and generalization in IQA\citep{song2023active,zhang2023blind}. Secondly, quality level recognition categorizes distorted images into levels~(e.g. high and bad ), based on human-perceptible semantic features. This natural language-based, range-oriented approach helps minimize errors in absolute scoring across different subjects~\citep{yang2020ttl}.

To address the problem (ii), we introduce a novel Perceptual Prior-based Feature Refinement Strategy for BIQA. Initially, we use pre-trained teacher pseudo features to establish a quality-aware denoising process based on text-conditioned DDIM. We then consider student features as noisy versions of teacher pseudo features. Through an adaptive noise alignment mechanism, we adaptively assess the noise level in each student feature and apply corresponding Gaussian noise, aligning these features with the teacher's predefined denoising path. During the reverse denoising, cross-attention with text conditions is conducted to precisely refine the quality-aware features.
Notably, our goal is to utilize the powerful denoising modeling capability of the diffusion model for feature refinement rather than learning the distribution of teacher pseudo features. Experiments show that we achieve superior performance with very few sampling steps (e.g. 5 iterations) and a lightweight diffusion model for denoising. 
We summarize the contributions of this work as follows:
\begin{itemize}
    \item 
    We make the first attempt to convert the challenges of BIQA to the diffusion problem for feature denoising. We introduce a novel PFD-IQA, which effectively filters out quality-irrelevant information from features.
    
    \item We propose a Perceptual Prior Discovery and Aggregation Module to identify perceptual features of distortion types and quality levels. By leveraging the correlation between perceptual prior and text embedding, we adaptively aggregate perceptual text prompts to guide the diffusion denoising process, which ensures attention to quality-aware features during the denoising process.

    \item We introduce a novel Perceptual Prior-based Feature Refinement Strategy for BIQA. Particularly,  we pre-define denoising trajectories of teacher pseudo-labels. Then, by employing an adaptive noise alignment module, we match the student noise features to predefined denoising trajectories and subsequently perform precise feature denoising based on the given prompt conditions.

\end{itemize}

\section{Related Work}
\subsection{BIQA with Deep Learning.}
The early BIQA method~\citep{ Liu_2017_ICCV,no,li2009natural} was based on the convolutional neural network~(CNN) thanks to its powerful feature expression ability.
The CNN-based BIQA methods ~\citep{zhang2018blind, hypernet} generally treated the IQA task as the downstream task of object recognition, following the standard pipeline of pre-training and fine-tuning. 
Such a strategy is useful as these pre-trained features share a certain degree of similarity with the quality-aware features of images~\citep{hypernet}.
Recently, Vision Transformer (ViT) based methods for Blind Image Quality Assessment (BIQA) have become popular due to their strong capability in modeling non-local perceptual features in images. There are two main types of architectures used: the hybrid transformer and the pure transformer. Existing ViT-based methods typically rely on the CLS token for assessing image quality. Originally designed for describing image content, like object recognition, the CLS token focuses on higher-level visual abstractions, such as semantics and spatial relationships of objects. Therefore, it is still a challenge to fully adapt these methods from classification tasks to image quality assessment (IQA) tasks, due to the abstract nature of classification features.
\begin{figure*}[t!]
  \centering
    \includegraphics[width=1\textwidth]{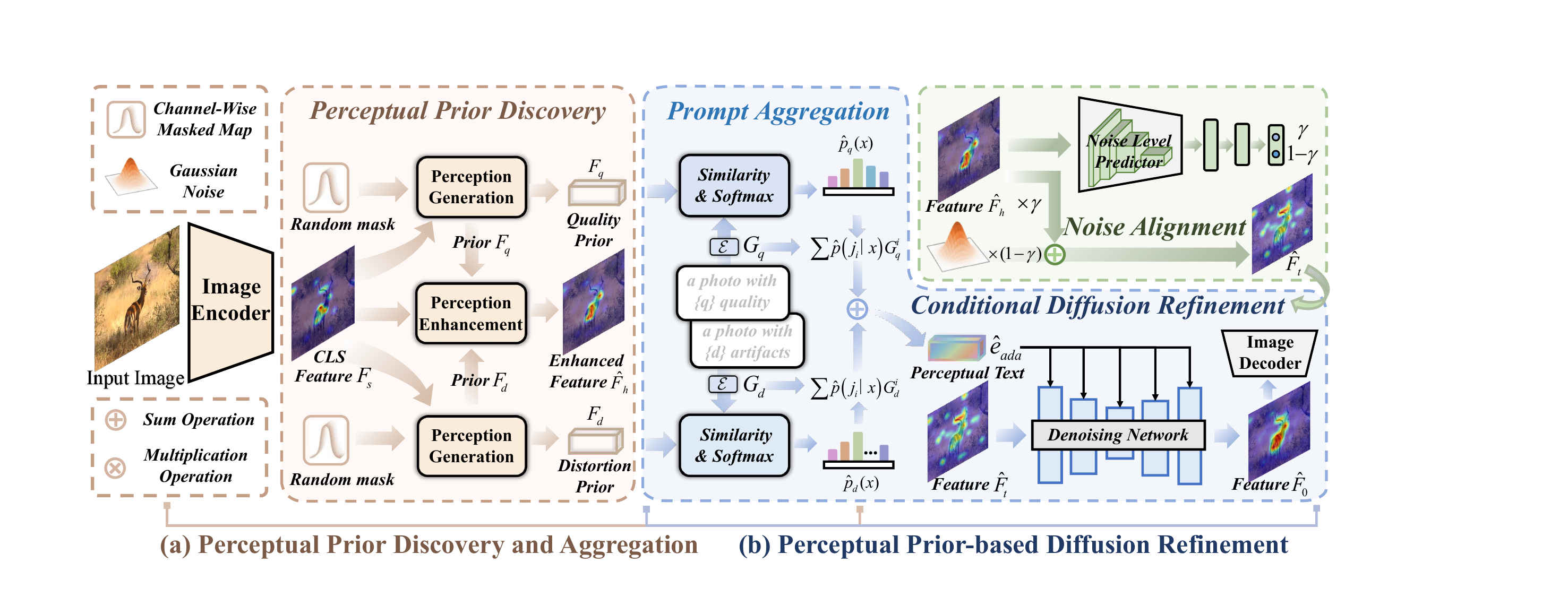}
  \caption{The overview of PFD-IQA, which consists of a teacher model used for creating pseudo-labels and a student model equipped with PDA and PDR modules. Specifically, we begin by developing a learning perceptual prior (Sec.~\ref{PDA}) through the random mask reconstruction process.  Subsequently,  we use the prior knowledge to aggregate text information as the condition to guide the feature-denoising process of the diffusion model and refine the features (Sec.~\ref{PDR}).
  }
  \label{framework}
\end{figure*}
\subsection{Diffusion Models.}
Diffusion models~\citep{rombach2022high,huang2023knowledge}, generally comprising a forward process for adding noise and a reverse process for denoising, gained popularity with Ho et al.'s introduction of the denoising diffusion probabilistic model. Building on this, methods like~\citep{rombach2022high} have integrated attention mechanisms into diffusion models, stabilizing them and producing high-quality images. To further extend diffusion models (dm) into mainstream computer vision tasks, latent representation learning methods based on dm have been proposed, including DiffusionDet for object detection~\citep{chen2023diffusiondet} and SegDiff for segmentation~\citep{amit2021segdiff}. However, diffusion models are seldom used for specific feature denoising. In this study, we treat the feature optimization process in IQA as an inverse denoising approximation and iteratively use diffusion models to enhance representations for accurate quality awareness. To the best of our knowledge, ours is the first work to introduce diffusion models into IQA for feature denoising.

\section{Methodology}
In the context of BIQA, we introduce common notations. Bold formatting is used to denote vectors (e.g., $\boldsymbol{x}$, $\boldsymbol{y}$), matrices (e.g., $\boldsymbol{X}$, $\boldsymbol{Y}$), and tensors. The training data consists of $D = \{x, y_{g}, y_{d}, y_{q}\}$, where $x$ is the labeled image with ground-truth scores $y_{g}$. $y_{d}$, and $y_{q}$ represent the distortion type and quality level pseudo-labels associated with the input image, respectively. Additionally, image embeddings $F$ and textual embeddings $G$ are denoted. The probability distribution of logits for the network is represented as $p$.
\subsection{Overview}
The paper introduces a model called the Perceptual Feature Diffusion model for Image Quality Assessment (PFD-IQA), which progressively refines quality-aware features. As depicted in Fig. \ref{framework}, PFD-IQA seamlessly integrates two main components: A Perceptual Prior Discovery and Aggregation (PDA) and A Perceptual Prior-based Diffusion Refinement Module (PDR). Initially, PFD-IQA inputs the given image $\boldsymbol{x}$ into a Vision Transformer (ViT) encoder \citep{ViT} to obtain a feature representation $\boldsymbol{F}_{s}$. Under the supervision of pseudo-labels for distortion types and quality levels, we use the PDA module to discover potential distortion priors $\boldsymbol{\hat{F}}_{d}$ and Perceptual priors $\boldsymbol{\hat{F}}_{q}$, which then adaptively aggregate perceptual text embeddings as conditions for the diffusion process~(Sec.~\ref{PDA}). Next, in the PDR module, these prior features are used to modulate $\boldsymbol{F}_{s}$ for feature enhancement to obtain $ \boldsymbol{\hat{F}_{h}}$. This is followed by matching it to a predefined noise level $\boldsymbol{\hat F}_{t}$ through an adaptive noise matching module $\epsilon$, and finally employing a lightweight feature denoising module to progressively denoise under the guidance of the perceptual text embeddings~(Sec.~\ref{PDR}). After the PDR module, a layer of transformer decoder is used to further interpret the denoised features for predicting the final quality score~\citep{DEIQT}. 
It is important to emphasize that pseudo-labels are only used for training.

\subsection{Perceptual Prior Discovery and Aggregation}~\label{PDA}
Considering the intricate nature of image distortions in the real world, the evaluation of image quality necessitates discriminative representations that can
distinguish different types of distortions~\citep{zhang2022contrastive}, as well as the degrees of degradation. To achieve this, an auxiliary task involving the classification of distortion types is introduced Which is designed to refine the differentiation among diverse distortion types, thereby providing nuanced information. Additionally, the quality levels classification task is further employed to offer a generalized classification that compensates for the uncertainty and error inherent in predicting absolute image quality scores.

\noindent\textbf{Perceptual Prior Discovery.}
In this context, two feature reconstructors denoted as $\mathcal {R(\cdot)}$ are trained to reconstruct the mentioned two prior features, respectively.
These reconstructors consist of two components: (1) a stochastic channel mask module and (2) a module for the feature reconstructions. Specifically, given an image $\boldsymbol{x}$ and its feature $\boldsymbol{F}_{s}$ that has been generated by a VIT encoder. The first step involves applying a channel-wise random mask $M_{c}$ to the channel dimension of this feature to obtain $\boldsymbol{F}_{m}$. 
\begin{equation}
\begin{split}
    M_{c} =   {\begin{cases}
    {0,}&{{ { if }}\ R_{c} < \beta }\\
    {1,}&{{ { Otherwise }}}
    \end{cases}}  , \quad
    {\boldsymbol{F}_{m}} =  {{f_{{ {align }}}} ( {{\boldsymbol{F}_{s}}}  ) \cdot {M_{c}}},
\end{split}
\end{equation}
where $R_{c}$ is a random number in (0, 1) and c are channel number of the feature. $\beta$ is a hyper-parameter that denotes the masked ratio and $f_{align}$ is a adaptation layer with 1×1 convolution. The random mask helps to train a more robust feature reconstructor~\citep{MGD}. Subsequently, we utilize the two feature reconstruction modules $\mathcal R(\cdot)$ to generate prior features. Each $\mathcal R(\cdot)$ consists of a sequence of operations including a 1×1 convolution ${W_{l1}}$, a Batch Normalization (BN) layer, and another 1×1 convolutional layer. ${W_{l2}}$.
\begin{equation}
\hat{\boldsymbol{F}}_{j} = \mathcal{R}(\boldsymbol{F}_{m}) = W_{l2, j} \cdot \left( \mathop{ReLU} \left( W_{l1, j}(\boldsymbol{F}_{m}) \right) \right),
\end{equation}
where, $j \in \{d, q\}$, $\hat{\boldsymbol{F}}_{d}$ and $\hat{\boldsymbol{F}}_{q}$ stands for distortion and quality level classification auxiliary tasks. These tasks are linked to the original image feature $\boldsymbol{F}_{s}$ and involve capturing different aspects of information.

To effectively supervise the auxiliary tasks related to quality level classification $Q$ and distortion classification $D$ for the discovery of potential prior features, we divide the tasks into five quality levels and eleven types of distortions following previous study~\citep{zhang2023blind}. 
As illustrated in Fig.~\ref{framework}, let \( D \) denote the set of image distortions, i.e., \( D = \{d_1, d_2, \ldots, d_K\} \), where \( d_i \) is an image distortion type, e.g., “noise”. Let \( Q \) denote the set of image quality levels, i.e., \( Q = \{q_1, q_2, \ldots, q_K\} \), where \( q_i \) is quality level, e.g., “bad”, and \( K \) is the number of distortions or quality levels we consider. 
The textural description prompt set is \( \mathcal{T}_d = \{T_d \mid T_d = \text{“a photo of with \{d\} artifact.”}, d \in D\} \) and \( \mathcal{T}_q = \{T_q \mid T_q = \text{“a photo of with $\{q\}$ quality.”}, q \in Q\} \). 
Given an image $x$, we compute the cosine similarity between image prior embedding \( \hat{F}_j \) and each prompt embedding $G_j = \mathcal{E}(T_j) \in \mathbb{R}^{K \times C}$ from text encoder $\mathcal{E}$ resulting in the logits output for the auxiliary tasks, namely $\boldsymbol{\hat p}_{d}$ and $\boldsymbol{\hat p}_{q}$, which the parameters of the text encoder is freeze:

\begin{equation}
   \boldsymbol {\hat p}_j(x) = {\text{logit}}(j|x) = \frac{\hat {\boldsymbol F} \cdot \boldsymbol G^\text{T}_j}{\|\hat {\boldsymbol F}_j\|_2 \|\boldsymbol G_j\|_2}.
   \label{logit_equation}
\end{equation}

To supervise the feature reconstruction module, we utilize the soft pseudo-labels $\boldsymbol{p}_{d}$ for distortion and $\boldsymbol{p}_{q}$ for quality, which are generated by the pre-trained teacher model. This guidance is accomplished by applying the KL divergence as follows:
\begin{equation}
   \mathcal{L}_{\mathrm{KL}} = \sum_{j \in \{d, q\}} \mathcal{L}_{K L}^j\left(\boldsymbol{p}_{j},  \hat{\boldsymbol{p}_{j}}\right) = \sum_{j \in \{d, q\}} \boldsymbol{p}_{x}\left(\boldsymbol{j}\right) \log \frac{\boldsymbol{p}_{j}\left(\boldsymbol{x}\right)}{\hat{\boldsymbol{p}_{j}}\left(\boldsymbol{x}\right)}
   \label{distill}
\end{equation}

\noindent\textbf{Perceptual Prompt Aggregation~(PPA).}
Psychological research suggests that humans prefer using natural language for qualitative rather than quantitative evaluations~\citep{hou2014blind}. In practice, this means qualitative descriptors like 'excellent' or 'bad' are often used to assess image quality. Building on this, we've developed an approach to automatically aggregate natural language prompts that qualitatively represent image quality perception. 
Specifically, we compute the logit for the distortion and quality levels, for each prompt, we can further get its \text{logit}$(j_i|I)$ by:
\begin{equation}
    \hat{p}(c_i|x) = \frac{\exp(\text{logit}(j_i|x))}{\sum_{k=1}^{K} \exp(\text{logit}(j_k|x))}, \\
\end{equation}
where \( j_i \) is the \( i \)-th element of \( T_j \). Next, we obtain the adaptive perceptual text embedding \( \hat{e}_{{ada}} \) via the following weighted aggregation:
\begin{equation}
    \hat{e}_{{ada}} = \sum_{i=1}^{K} \hat{p}(j_i|x) G^i_J,\quad j \in \{d, q\}
\end{equation}

It is worth noting that $\hat{e}_{{ada}}$ is capable of effectively representing the multi-distortion mixture information in real distorted images as soft label weightings. This approach is more informative compared to the hard label method, which relies solely on a single image-text pair.

\begin{figure}[t!]
  \centering
    \includegraphics[width=0.48\textwidth]{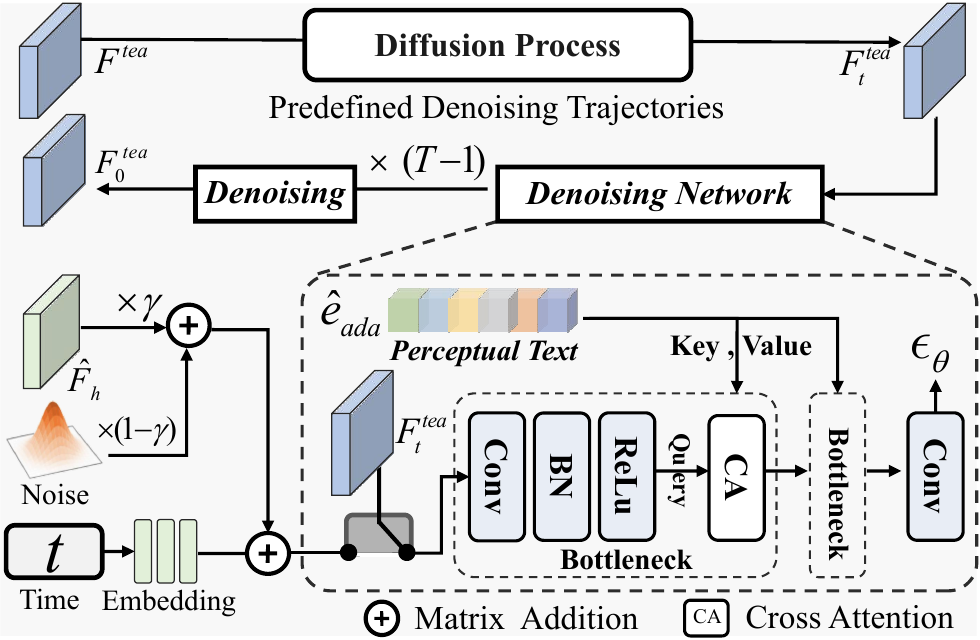}
  \caption{
The predefined denoising trajectory starts with a teacher pseudo-feature label for forward diffusion. During each reverse denoising phase, image and text information are fused to accurately predict the noise in the features. For student denoising, the noise level matched by the noise alignment mechanism is used as the input for noise prediction.}
  \label{submodel}
\end{figure}

\subsection{Perceptual Prior-based Diffusion Refinement}~\label{PDR}
In this section, we introduce our Perceptual Prior Fusion~(PPF) Module and Perceptual Prior-based Diffusion Refinement Module~(PDR), as well as discuss how to automatically aggregate perceptual text embeddings that can be used as conditions to guide feature denoising.

\noindent\textbf{Perceptual Prior Enhancement.}
Due to the primary emphasis on global semantic features in pre-trained models, there exists a gap in capturing quality-aware information across different granularities. To address this, we propose the integration of perceptual prior information to enhance feature representations. 
Specifically, we introduce the Perceptual Prior Fusion module (PPF), which is designed to merge both distortion perception and quality degradation perception into the framework. The proposed PPF Module operates sequentially on normalized features, incorporating additional convolutions and SiLU layers~\citep{elfwing2018sigmoid} to facilitate the fusion of features across different granularities. In the implementation, we first apply a two-dimensional scaling modulation to the normalized feature norm $\boldsymbol{F}_{s}$ and then employ two convolutional transformations modulate the normalized feature $\boldsymbol{F}_{s}$ with scaling and shifting parameters from additive features $\boldsymbol {\hat{F}}_{dq}$, resulting in the feature representation $\boldsymbol {\hat{F}}_{h}$:
\begin{equation}
 \hat{\boldsymbol F}_{h} =(\text{conv}(\boldsymbol{ \hat{F}}_{dq}) \times \text{norm}(\boldsymbol{F}_{s}) + \text{conv}(\boldsymbol{ \hat{F}}_{dq})) + \boldsymbol{F}_{s}.
\end{equation}


\noindent\textbf{Predefined Conditional Denoising Trajectories.}
The proposed PFD-IQA iteratively optimizes the feature $ \boldsymbol{\hat{F}_{h}}$ to attain accurate and quality-aware representations. This process can be conceptualized as an approximation of the inverse feature denoising procedure. However, the features representing the ground truth are often unknown. Therefore, we introduce features $\boldsymbol{F}^{tea}$ generated by a pre-trained teacher as pseudo-ground truth to pre-construct a denoising trajectory of quality-aware features. As depicted in Fig.~\ref{submodel}, for the forward diffusion process, $\boldsymbol{F}^{tea}_t$ is a linear combination of the initial data $\boldsymbol{F}^{tea}$ and the noise variable $\epsilon_t$.
\begin{equation}
\boldsymbol{F}^{tea}_t = \sqrt{\bar\alpha_t} \boldsymbol{F}^{tea} + \sqrt{1 - \bar\alpha_t} \epsilon_t.
\end{equation}

The parameter \(\bar\alpha_t\) is defined as \(\bar\alpha_t:= \prod_{s=0}^t \alpha_s = \prod_{s=0}^t (1 - \beta_s)\), offering a method to directly sample $\boldsymbol{F}^{tea}_t$ at any time step using a noise variance schedule denoted by \(\beta\)~\citep{ho2020denoising}.
During training, a neural network ${\epsilon}_{\theta}(\boldsymbol{F}^{tea}_t, \boldsymbol{\hat{e}}^{tea}_{{ada}},t)$ conditioned on perceptual text $\boldsymbol{\hat{e}}^{tea}_{{ada}}$ is trained to predict the noise $\boldsymbol{\epsilon}_t \in \mathcal{N}(\mathbf{0}, \boldsymbol{I})$ by minimizing the ${\ell_2}$ loss, i.e.,
\begin{equation}
\mathcal{L}_{ldm} = \| {\epsilon}_t - {\epsilon}_{\theta}(\boldsymbol{F}^{tea}_t, \boldsymbol{\hat{e}}^{tea}_{{ada}}, t)\|_{2}^{2},
\end{equation}

\noindent\textbf{Adaptive Noise-Level Alignment~(ANA).}
We treat the feature representations extracted by students according to the fine-tuning paradigm as noisy versions of the teacher's quality-aware features.
However, the extent of noise that signifies the dissimilarity between the teacher and student features remains elusive and may exhibit variability across distinct training instances. As a result, identifying the optimal initial time step to initiate the diffusion process presents a challenging task. To overcome this, we introduce an Adaptive Noise Matching Module to match the noise level of student features with a predefined noise level.

As depicted in Fig.~\ref{framework}, we develop a Noise-level predictor using a straightforward convolutional module aimed at learning a weight $\gamma$ to combine the fusion feature $\boldsymbol {\hat{F}}_{h}$ of the student with Gaussian noise, resulting in $\boldsymbol{\hat F}_t$ that aligns with $\boldsymbol{F}_t$. This weight ensures that the student's outputs are harmonized with the noise level corresponding to the initial time step $t$. Consequently, the initial noisy feature involved in the denoising process is altered subsequently:
\begin{equation}
\boldsymbol{\hat F}_{t} = \gamma \odot \boldsymbol{\hat 
 F}_{h} + (1 - \gamma) \odot \mathcal{N}(0, 1)
\label{noise}
\end{equation}

\noindent\textbf{Lightweight Architecture.}
Considering the huge dimension of transformers, performing the denoising process on features during training requires considerable iterations, which may result in a huge computational load. To address this issue, this paper proposes a lightweight diffusion model $\epsilon_{\theta}(\cdot)$ as an alternative to the U-net architecture, as shown in Fig.~\ref{submodel}. It consists of two bottleneck blocks from ResNet \citep{resnet} and a 1×1 convolution. In addition, A cross-attention layer~\citep{rombach2022high} is added after each bottleneck block to aggregate the text and image features. We empirically find that this lightweight network is capable of effective noise removal with less than 5 sampling iterations which is more than 200$\times$ faster sampling speed compared to the DDPM.

During the sampling process, with the initial noise $\boldsymbol{F}_t$ obtained in Equ.~\ref{noise}, the trained network is employed for iterative denoising to reconstruct the feature $\boldsymbol{\hat F}_0$:
\begin{equation}
p_\theta\left(\boldsymbol{\hat F}_{t-1} \mid \boldsymbol{\hat F}_t\right):=\mathcal{N}\left(\boldsymbol{\hat F}_{t-1} ; \epsilon_\theta\left(\boldsymbol{\hat F}_t,\boldsymbol{\hat{e}}_{{ada}}, t\right), \sigma_t^2 \boldsymbol{I}\right)
\end{equation}
Subsequently, we employ the features $\boldsymbol{F}^{tea}$ derived from the pseudo-labels generated by the pre-trained teacher to supervise the denoising procedure using MSE loss which ensures the stability of the feature denoising process.
\begin{equation}
\mathcal{L}_{fea} = \| {\boldsymbol{\hat F}_0 - \boldsymbol{F}^{tea})}\|_{2}^{2}
\label{fea_distill}
\end{equation}
To sum up, the overall loss at the training stage is described as follows:
\begin{equation}
\mathcal{L} = \lambda_1\mathcal{L}_{KL} + \lambda_2\mathcal{L}_{ldm}+ \lambda_3\mathcal{L}_{fea} +  \| {\boldsymbol{\hat y} - \boldsymbol{y}_g)}\|_1
\end{equation}
Here, $\hat y$ represents the predicted score of image $x$ based on the denoised feature obtained from the transformer decoder. $\boldsymbol{y}_g$ stands for the ground truth corresponding to the image $x$. The notation ${\|\cdot\|}_1$ denotes the ${\ell_1}$ regression loss. In this paper, We simply set $\lambda_1 = 0.5$, $\lambda_2 = 1$, and $\lambda_3 = 0.01$ in all experiments.

\section{Experiments}
\vspace{-0.8cm}
\begin{center} 
\begin{table*}[t]
\setlength\tabcolsep{0.7pt}
    \centering
    \resizebox{1\textwidth}{!}{
        \begin{tabular}{lcccccccc||cccccccc}
      \toprule[1.5pt] & \multicolumn{2}{c}{ LIVE } & \multicolumn{2}{c}{ CSIQ } & \multicolumn{2}{c}{ TID2013 } & \multicolumn{2}{c||}{ KADID } & \multicolumn{2}{c}{ LIVEC } & \multicolumn{2}{c}{ KonIQ } & \multicolumn{2}{c}{ LIVEFB } & \multicolumn{2}{c}{ SPAQ } \\
      \cmidrule{2-17}    Method & \multicolumn{1}{c}{PLCC} & \multicolumn{1}{c}{SRCC} & \multicolumn{1}{c}{PLCC} & \multicolumn{1}{c}{SRCC} & \multicolumn{1}{c}{PLCC} & \multicolumn{1}{c}{SRCC}& \multicolumn{1}{c}{PLCC} & \multicolumn{1}{c||}{SRCC}& \multicolumn{1}{c}{PLCC} & \multicolumn{1}{c}{SRCC}& \multicolumn{1}{c}{PLCC} & \multicolumn{1}{c}{SRCC}& \multicolumn{1}{c}{PLCC} & \multicolumn{1}{c}{SRCC}& \multicolumn{1}{c}{PLCC} & \multicolumn{1}{c}{SRCC}\\
    \midrule
    DIIVINE ~\citep{saad2012blind} & 0.908 & 0.892 & 0.776 & 0.804 & 0.567 & 0.643 & 0.435 & 0.413 & 0.591 & 0.588 & 0.558 & 0.546 & 0.187 & 0.092 & 0.600 & 0.599 \\
    BRISQUE ~\citep{BRISQUE} & 0.944 & 0.929 & 0.748 & 0.812 & 0.571 & 0.626 & 0.567 & 0.528 & 0.629 & 0.629 & 0.685 & 0.681 & 0.341 & 0.303 & 0.817 & 0.809 \\
    ILNIQE ~\citep{ILNIQE} & 0.906 & 0.902 & 0.865 & 0.822 & 0.648 & 0.521 & 0.558 & 0.534 & 0.508 & 0.508 & 0.537 & 0.523 & 0.332 & 0.294 & 0.712 & 0.713 \\
    BIECON ~\citep{BIECON} & 0.961 & 0.958 & 0.823 & 0.815 & 0.762 & 0.717 & 0.648 & 0.623 & 0.613 & 0.613 & 0.654 & 0.651 & 0.428 & 0.407 & {-} & {-} \\
    MEON ~\citep{MEON} & 0.955 & 0.951 & 0.864 & 0.852 & 0.824 & 0.808 & 0.691 & 0.604 & 0.710  & 0.697 & 0.628 & 0.611 & 0.394 & 0.365 & {-} & {-} \\
    WaDIQaM ~\citep{bosse2017deep} & 0.955 & 0.960  & 0.844 & 0.852 & 0.855 & 0.835 & 0.752 & 0.739 & 0.671 & 0.682 & 0.807 & 0.804 & 0.467 & 0.455 & {-} & {-} \\
    DBCNN ~\citep{zhang2018blind} & 0.971 & 0.968 & {0.959} & {0.946} & 0.865 & 0.816 & 0.856 & 0.851 & 0.869 & 0.851 & 0.884 & 0.875 & 0.551 & 0.545 & 0.915 & 0.911 \\
    TIQA ~\citep{TIQA} & 0.965 & 0.949 & 0.838 & 0.825 & 0.858 & 0.846 & 0.855 & 0.85  & 0.861 & 0.845 & 0.903 & 0.892 & 0.581 & 0.541 & {-} & {-} \\
    MetaIQA ~\citep{zhu2020metaiqa} & 0.959 & 0.960  & 0.908 & 0.899 & 0.868 & 0.856 & 0.775 & 0.762 & 0.802 & 0.835 & 0.856 & 0.887 & 0.507 & 0.54  & {-} & {-} \\
    P2P-BM ~\citep{ying2020patches} & 0.958 & 0.959 & 0.902 & 0.899 & 0.856 & 0.862 & 0.849 & 0.84  & 0.842 & 0.844 & 0.885 & 0.872 & 0.598 & 0.526 & {-} & {-} \\
    HyperIQA ~\citep{hypernet} & 0.966 & 0.962 & 0.942 & 0.923 & 0.858 & 0.840  & 0.845 & 0.852 & 0.882 & 0.859 & 0.917 & 0.906 & 0.602 & 0.544 & 0.915 & 0.911 \\
    TReS  ~\citep{TReS} & 0.968 & 0.969 & 0.942 & 0.922 & 0.883 & 0.863 & 0.858 & 0.859 & 0.877 & 0.846 & {0.928} & 0.915 & 0.625 & 0.554 & {-} & {-} \\
    MUSIQ ~\citep{ke2021musiq} & 0.911 & 0.940  & 0.893 & 0.871 & 0.815 & 0.773 & 0.872 & 0.875 & 0.746 & 0.702 & {0.928} & {0.916} & {0.661} & {0.566} & {0.921} & {0.918} \\
    DACNN ~\citep{pan2022dacnn} & {0.980}  & {0.978} & {0.957} & {0.943} & {0.889} & {0.871} & \underline{0.905} & \underline{0.905} & {0.884} & {0.866} & 0.912 & 0.901 & {-} & {-} & {0.921} & 0.915 \\
    DEIQT ~\citep{DEIQT} & \underline{0.982} & \underline{0.980} & \underline{0.963} & \underline{0.946} & \underline{0.908} & \underline{0.892} & 0.887 & 0.889 & \underline{0.894} & \underline{0.875} & \underline{0.934} & \underline{0.921} & \underline{0.663} & \underline{0.571} & \underline{0.923} & \underline{0.919} \\
    \midrule
    \rowcolor{gray!20} PFD-IQA (ours) & \textbf{0.985} & \textbf{0.985} & \textbf{0.972} & \textbf{0.962} & \textbf{0.937} & \textbf{0.924} & \textbf{0.935} & \textbf{0.931} & \textbf{0.922} & \textbf{0.906} & \textbf{0.945} & \textbf{0.930} & \textbf{0.667} & \textbf{0.572} & \textbf{0.925} & \textbf{0.922} \\
    \bottomrule
    \end{tabular}}
  \caption{Performance comparison measured by averages of SRCC and PLCC, where bold entries indicate the best results, \underline{underlines} indicate the second-best.} 
  \vspace{-0.2cm}
  \label{performance}
\end{table*}
\end{center}

\begin{table}[t]
\small
\setlength\tabcolsep{4pt}
  \centering
    \begin{tabular}{ccccccc}
    \toprule
    Training & \multicolumn{2}{c}{  LIVEFB } & \multicolumn{1}{c}{LIVEC} & \multicolumn{1}{c}{KonIQ} & \multicolumn{1}{c}{LIVE} & \multicolumn{1}{c}{CSIQ} \\
    \midrule
    Testing & \multicolumn{1}{c}{KonIQ} & \multicolumn{1}{c}{LIVEC} & \multicolumn{1}{c}{KonIQ} & \multicolumn{1}{c}{LIVEC} & \multicolumn{1}{c}{CSIQ} & \multicolumn{1}{c}{LIVE} \\
    \midrule
    DBCNN & 0.716 & 0.724 & 0.754 & 0.755 & 0.758 & 0.877 \\
    P2P-BM & 0.755 & 0.738 & {0.740}  & 0.770  & 0.712 & {-} \\
    HyperIQA & \underline{0.758} & 0.735 & \underline{0.772} & 0.785 & 0.744 & {0.926} \\
    TReS  & 0.713 & 0.740  & 0.733 & {0.786} & {0.761} & {-} \\
    DEIQT & 0.733 & \underline{0.781} & 0.744 & \underline{0.794} & \underline{0.781} & \underline{0.932} \\
    \midrule
    \rowcolor{gray!20} PFD-IQA   & \textbf{0.775} & \textbf{0.783} & \textbf{0.796}  & \textbf{0.818} & \textbf{0.817} & \textbf{0.942} \\
    \bottomrule
    \end{tabular}%
  \caption{SRCC on the cross datasets validation. The best results are highlighted in bold, second-best is \underline{underlined}.}
  \label{cross}%
\end{table}%

\subsection{Benchmark Datasets and Evaluation Protocols}
We evaluate the performance of the proposed PFD-IQA on eight typical BIQA datasets, including four synthetic datasets of LIVE ~\citep{sheikh2006statistical}, CSIQ ~\citep{larson2010most}, TID2013 ~\citep{ponomarenko2015image}, KADID ~\citep{lin2019kadid}, and four authentic datasets of
LIVEC~\citep{ghadiyaram2015massive}
KONIQ~\citep{hosu2020koniq}, LIVEFB ~\citep{ying2020patches}, SPAQ ~\citep{fang2020perceptual}. Specifically, for the authentic dataset, LIVEC contains 1162 images from different mobile devices and photographers. SPAQ comprises 11,125 photos from 66 smartphones. KonIQ-10k includes 10,073 images from public sources, while LIVEFB is the largest real-world dataset to date, with 39,810 images. The synthetic datasets involve original images distorted artificially using methods like JPEG compression and Gaussian blur. LIVE and CSIQ have 779 and 866 synthetically distorted images, respectively, with five and six distortion types each. TID2013 and KADID include 3000 and 10,125 synthetically distorted images, respectively, spanning 24 and 25 distortion types.

In our experiments, we employ two widely used metrics: Spearman's Rank Correlation Coefficient (SRCC) and Pearson's Linear Correlation Coefficient (PLCC). These metrics evaluate prediction monotonicity and accuracy, respectively.
\subsection{Implementation Details}
For the student network, we follow the typical training strategy of randomly cropping the input image into 10 image patches with a resolution of $224 \times  224$. Each image patch is then reshaped as a sequence of patches with patch size p = 16 and the dimension of input tokens as in $D = 384$. We create the Transformer encoder based on the ViT-B proposed in DeiT III ~\citep{touvron2022deit}. The encoder depth is set to 12 and the number of heads h = 12. For Decoder, the depth is set to one. Our model is trained for 9 epochs. The learning rate is set to ${2 \times  10^{-4}}$ with a decay factor of 10 every 3 epochs. The batch size depends on the size of the dataset, which is 16 and 128 for LIVEC and KonIQ, respectively. For each dataset, 80\% of the images are used for training and the remaining 20\% of the images are used for testing. We repeat this process 10 times to mitigate performance bias and report the average of SRCC and PLCC. For the pre-trained teacher network, We adopt ViT-B/16~\citep{radford2021learning} as the visual encoder and text encoder. The re-training hyperparameter Settings are consistent with \citep{zhang2023blind}.

\begin{center} 
\begin{table*}[t]
\setlength\tabcolsep{0.7pt}
    \centering
    \resizebox{1\textwidth}{!}{
   \begin{tabular}{c|c|c|ccccccccccc}
    \toprule 
    \multirow{2}{*}{ Index } & \multirow{2}{*}{$\boldsymbol {PDA}$} & \multirow{2}{*}{$\boldsymbol {PDR}$} & \multicolumn{2}{c}{LIVE} & \multicolumn{2}{c}{CSIQ}& \multicolumn{2}{c}{TID2013} & \multicolumn{2}{c}{KADID} & \multicolumn{2}{c}{LIVEC} & \multirow{2}{*}{Avg.}
    \\\cmidrule(r){4-5}\cmidrule(r){6-7}\cmidrule(r){8-9}\cmidrule(r){10-11}\cmidrule(r){12-13}
     & & & \multicolumn{1}{c}{PLCC} & \multicolumn{1}{c}{SRCC}  & \multicolumn{1}{c}{PLCC}  & \multicolumn{1}{c}{SRCC} & \multicolumn{1}{c}{PLCC}  & \multicolumn{1}{c}{SRCC} & \multicolumn{1}{c}{PLCC}  & \multicolumn{1}{c}{SRCC} & \multicolumn{1}{c}{PLCC}  & \multicolumn{1}{c}{SRCC} & \\
    \midrule
    \textbf{\textit{a)}}       &   &    & 0.966 & 0.964 & 0.952 & 0.935 & 0.899 & 0.888 & 0.878 & 0.884 & 0.881 & 0.863 & 0.911       \\
    \textbf{\textit{b)}}       & \CheckmarkBold  &     & 0.984 & 0.981 & 0.963 & 0.954 & 0.927 & 0.915 & 0.925 & 0.925 & 0.911 & 0.895 & 0.938        \\
    \textbf{\textit{c)}}       &   &  \CheckmarkBold   & 0.983 & 0.982 & 0.968 & 0.959 & 0.910 & 0.890  & 0.918 & 0.919 & 0.916 & 0.897 & 0.934       \\
    \rowcolor{gray!20} \textbf{\textit{d)}}       & \CheckmarkBold  &  \CheckmarkBold   & $\textbf{0.985}_{+1.9\%}$ & $\textbf{0.985}_{+2.1\%}$ & $\textbf{0.972}_{+2.0\%}$ & $\textbf{0.962}_{+2.7\%}$ & $\textbf{0.937}_{+3.8\%}$ & $\textbf{0.924}_{+3.6\%}$ & $\textbf{0.935}_{+6.0\%}$ & $\textbf{0.931}_{+4.7\%}$ & $\textbf{0.922}_{+4.1\%}$ & $\textbf{0.906}_{+4.3\%}$ & $\textbf{0.946}_{+3.5\%}$       \\
    \bottomrule
    \end{tabular}}
      \caption{Ablation experiments on LIVE, CSIQ, TID2013, KADAD and LIVEC datasets. Here, $\boldsymbol {PDA}$ and $\boldsymbol {PDR}$ refer to the Perceptual Prior Discovery and Aggregation module and Diffusion Refinement module, where bold entries indicate the best result.}
      \vspace{-0.2cm}
    \label{ablation}
\end{table*}
\end{center}

\subsection{Overall Prediction Performance Comparison}
For competing models,  we either directly adopt the publicly available implementations, or re-train them on our datasets with the training codes provided by the respective authors. Tab. \ref{performance} reports the comparison results between the proposed PFD-IQA and 14 state-of-the-art BIQA methods, including hand-crafted feature-based BIQA methods, such as ILNIQE (Zhang, Zhang, and Bovik 2015) and BRISQUE (Mittal, Moorthy, and Bovik 2012), and deep learning-based methods, i.e., MUSIQ ~\citep{ke2021musiq} and MetaIQA ~\citep{zhu2020metaiqa}. It is observed from these eight datasets that PFD-IQA achieves superior performance over all other methods across the 8 datasets. Since the images on these 8 datasets cover various image content and distortion types, it is very challenging to consistently achieve leading performance on all these datasets. Accordingly, these observations confirm the effectiveness and superiority of PFD-IQA in characterizing image quality.

\subsection{Generalization Capability Validation}
We further evaluate the generalization ability of PFD-IQA by a cross-dataset validation approach, where the BIQA model is trained on one dataset and then tested on the others without any fine-tuning or parameter adaptation. Tab. \ref{cross} reports the experimental results of SRCC averages on the five datasets. As observed, PFD-IQA achieves the best performance on six cross-datasets, achieving clear performance gains on the LIVEC dataset and competitive performance on the KonIQ dataset. These results strongly verify the generalization ability of PFD-IQA.

\begin{figure}[t!]
\centering{\includegraphics[width=0.47\textwidth]{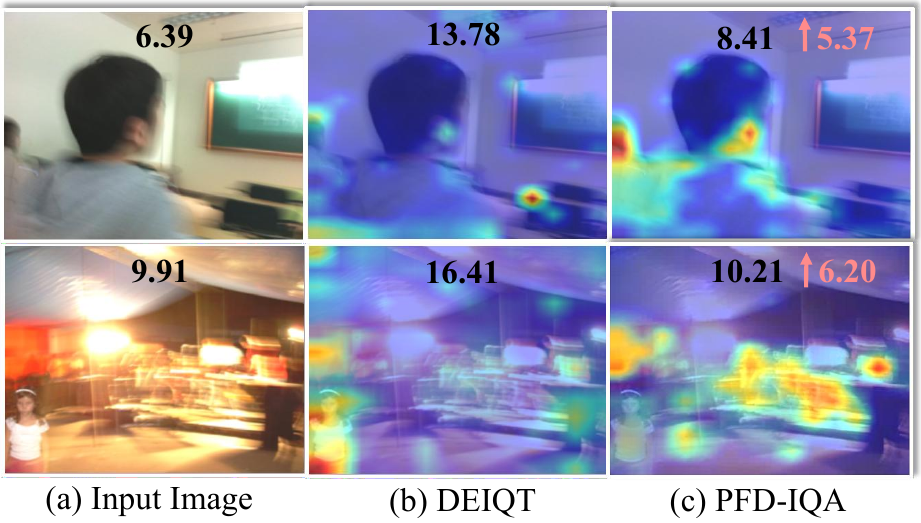}}
\caption{Visualization of the feature of DEIQT and PFD-IQA.}
 \label{visualize}
\end{figure}

\subsection{Qualitative Analysis}
\noindent \textbf{Visualization of activation map}
We employ GradCAM~\citep{grad} to visualize the feature attention map, as shown in Fig.~\ref{visualize}.
Our findings indicate that PFD-IQA effectively focuses on the quality degradation areas, while the DEIQT\citep{DEIQT}~(the second best in Tab.~\ref{performance}) 
overly relies on semantics and even focuses on less important for the image quality(e.g., in the second line, DEIQT incorrectly focuses on the little girl in the bottom left, while ignoring the overexposure and blurriness in the center of the picture.).
Furthermore, Fig.~\ref{visualize} presents a comparison of quality predictions between our proposed PFD-IQA and the DEIQT.
PFD-IQA consistently outperforms the DEIQT across all levels of image quality assessment, particularly demonstrating significant improvement for moderately distorted images, refer to the supplementary material for more visualizations.


\begin{center} 
\begin{table}[t]
\setlength\tabcolsep{5.2pt}
    \centering
    \resizebox{0.47\textwidth}{!}{
   \begin{tabular}{c|c|c|cccc|c}
    \toprule 
    \multirow{2}{*}{ Index } & \multirow{2}{*}{$\boldsymbol {ANA}$} & \multirow{2}{*}{$\boldsymbol {PPA}$} & \multicolumn{2}{c}{LIVEC} & \multicolumn{2}{c}{KonIQ} & \multirow{2}{*}{Avg. Std.}
    \\\cmidrule(r){4-5}\cmidrule(r){6-7}
     & & & \multicolumn{1}{c}{PLCC} & \multicolumn{1}{c}{SRCC}  & \multicolumn{1}{c}{PLCC}  & \multicolumn{1}{c}{SRCC} & \\
    \midrule
    \textbf{\textit{a)}}       &   &    & 0.881 & 0.863 & 0.929 & 0.914 & ±0.011      \\
    \textbf{\textit{b)}}       & \CheckmarkBold  & & 0.914 & 0.896 & 0.938 & 0.928 & ±0.005       \\
    \textbf{\textit{c)}}       &   &  \CheckmarkBold   & 0.918 & 0.900 & 0.940 & 0.928 & ±0.008     \\
    \rowcolor{gray!20} \textbf{\textit{d)}}       & \CheckmarkBold  &  \CheckmarkBold   & $\textbf{0.922}$ & $\textbf{0.906}$ & $\textbf{0.945}$ & $\textbf{0.930}$ &  \textbf{±0.004}    \\
    \bottomrule
    \end{tabular}}
  \caption{Ablation experiments about PDR component on LIVEC and KonIQ datasets. Bold entries indicate the best performance.} 
  \label{ablation_sub}
\end{table}
\end{center}

\vspace{-0.4cm}
\subsection{Ablation Study}
This section presents an ablation experiment, and the results are shown in Tab.~\ref{ablation}. The experiment is focused on the examination of two main modules: the Perceptual Prior Discovery and Aggregation (PDA) module and the Perceptual Prior-based Diffusion Refinement (PDR) module. 

\noindent \textbf{Perceptual Prior Discovery and Aggregation.}
When exclusively incorporating the PDA module (referred to as scenario \emph{b)}), discernible enhancements are observed in the PLCC (Pearson Linear Correlation Coefficient) ranging from 0.8\% to 4.7\% across distinct datasets. This outcome underscores the module's efficacy in augmenting the model's awareness of quality through the assimilation of supplementary quality-related knowledge. A similar positive impact is observed in the SRCC, manifesting as an improvement ranging from 0.7\% to 4.1\% across different datasets.

\noindent \textbf{Feature Diffusion Refinement Module.}
Upon sole integration of the PDR module (referred to as scenario \emph{c)}), discernible improvements are discerned in the PLCC, with advancements ranging from 0.9\% to 4.0\% across diverse datasets. Correspondingly, there is a notable enhancement in SRCC, with gains varying from 0.8\% to 3.5\% across these datasets. This observation suggests that the PDR module possesses the capacity to effectively optimize features, even in the absence of the perceptual text prompts condition.
Additionally, combining the PDR and PDA modules (scenario \emph{d)}) notably enhances PLCC (3.8 to 6.0\%) and SRCC (3.6 to 4.7\%). This highlights their synergistic effect, substantially improving PFD-IQA's robustness and accuracy.

\noindent \textbf{ANA and PPA module.}
As shown in Tab.~\ref{ablation_sub}, the PPA module alone enhances performance by up to 3.4\% over baseline, leveraging perceptual text embedding cues, though lacking in stability. The ANA module, on the other hand, significantly lowers the model's standard deviation and offers competitive performance improvements. When combined, ANA aligns features with predefined denoising modules, reducing randomness, while PPA precisely refines perceptual features via image-text interaction. Consequently, the synergistic cooperation of both modules helps achieve SOTA performance.

\noindent \textbf{Number of Sampling Iterations.} 
In our study, we employ DDIM~\citep{ddim} for acceleration. 
The experiments emphasize how different sampling numbers impact performance. Tab.~\ref{ablation_sample} demonstrates that single-step denoising significantly outperforms the baseline. We find that an iteration of 5 is adequate for effective performance in our approach. Therefore, this method is adopted in all experiments to balance efficiency and accuracy.

\begin{center}
\begin{table}[t]
\setlength\tabcolsep{6.2pt}
    \centering
    \resizebox{0.47\textwidth}{!}{
   \begin{tabular}{c|cccc|c}
    \toprule 
    \multirow{2}{*}{Sampling Number T} & \multicolumn{2}{c}{LIVE} & \multicolumn{2}{c}{LIVEC} & \multirow{2}{*}{Avg.}
    \\\cmidrule(r){2-3}\cmidrule(r){4-5}
     & \multicolumn{1}{c}{PLCC} & \multicolumn{1}{c}{SRCC}  & \multicolumn{1}{c}{PLCC}  & \multicolumn{1}{c}{SRCC} & \\
    \midrule
    1       & 0.982 & 0.980 & 0.913 & 0.890 & 0.941     \\
    3       & 0.984 & 0.982 & 0.918 & 0.899 & 0.945        \\
    \rowcolor{gray!20} 5       & \textbf{0.985} & \textbf{0.985} & \textbf{0.922} & \textbf{0.906} & 0.950      \\
    10       & 0.983 & 0.983 & 0.921 & 0.901 & 0.947     \\
    \bottomrule
    \end{tabular}}
  \caption{Ablation experiments about the number of sampling iterations. Bold entries indicate the best performance.} 
  \label{ablation_sample}
\end{table}
\end{center}

\vspace{-0.6cm}
\section{Conclusion} 
In conclusion, our study introduces the PFD-IQA, a pioneering Blind Image Quality Assessment framework leveraging the diffusion model's noise removal capabilities. Addressing key challenges in BIQA, PFD-IQA features two novel modules: the Perceptual Prior Discovery and Aggregation module for improved feature preservation and noise elimination, and the Perceptual Prior-based Feature Refinement Strategy for defining denoising trajectories in the absence of explicit benchmarks. These innovations, combining text prompts with perceptual priors and employing an adaptive noise alignment mechanism, enable PFD-IQA to refine quality-aware features with precision. Our experiments demonstrate that PFD-IQA achieves exceptional performance with minimal sampling steps and a lightweight model, marking a significant advancement in the application of diffusion models to image quality assessment.

\bibliographystyle{named}
\bibliography{ijcai24}

\begin{thebibliography}{}

\bibitem[\protect\citeauthoryear{Amit \bgroup \em et al.\egroup }{2021}]{amit2021segdiff}
Tomer Amit, Tal Shaharbany, Eliya Nachmani, and Lior Wolf.
\newblock Segdiff: Image segmentation with diffusion probabilistic models.
\newblock {\em arXiv preprint arXiv:2112.00390}, 2021.

\bibitem[\protect\citeauthoryear{Banham and Katsaggelos}{1997}]{banham1997digital}
Mark~R Banham and Aggelos~K Katsaggelos.
\newblock Digital image restoration.
\newblock {\em IEEE signal processing magazine}, 14(2):24--41, 1997.

\bibitem[\protect\citeauthoryear{Bosse \bgroup \em et al.\egroup }{2017}]{bosse2017deep}
Sebastian Bosse, Dominique Maniry, Klaus-Robert M{\"u}ller, Thomas Wiegand, and Wojciech Samek.
\newblock Deep neural networks for no-reference and full-reference image quality assessment.
\newblock {\em IEEE Transactions on image processing}, 27(1):206--219, 2017.

\bibitem[\protect\citeauthoryear{Chen \bgroup \em et al.\egroup }{2023}]{chen2023diffusiondet}
Shoufa Chen, Peize Sun, Yibing Song, and Ping Luo.
\newblock Diffusiondet: Diffusion model for object detection.
\newblock In {\em Proceedings of the IEEE/CVF International Conference on Computer Vision}, pages 19830--19843, 2023.

\bibitem[\protect\citeauthoryear{Dong \bgroup \em et al.\egroup }{2015}]{dong2015image}
Chao Dong, Chen~Change Loy, Kaiming He, and Xiaoou Tang.
\newblock Image super-resolution using deep convolutional networks.
\newblock {\em IEEE transactions on pattern analysis and machine intelligence}, 38(2):295--307, 2015.

\bibitem[\protect\citeauthoryear{Dosovitskiy \bgroup \em et al.\egroup }{2021}]{ViT}
Alexey Dosovitskiy, Lucas Beyer, Alexander Kolesnikov, Dirk Weissenborn, Xiaohua Zhai, Thomas Unterthiner, Mostafa Dehghani, Matthias Minderer, Georg Heigold, Sylvain Gelly, Jakob Uszkoreit, and Neil Houlsby.
\newblock An image is worth 16x16 words: Transformers for image recognition at scale.
\newblock In {\em International Conference on Learning Representations}, 2021.

\bibitem[\protect\citeauthoryear{Elfwing \bgroup \em et al.\egroup }{2018}]{elfwing2018sigmoid}
Stefan Elfwing, Eiji Uchibe, and Kenji Doya.
\newblock Sigmoid-weighted linear units for neural network function approximation in reinforcement learning.
\newblock {\em Neural networks}, 107:3--11, 2018.

\bibitem[\protect\citeauthoryear{Fang \bgroup \em et al.\egroup }{2020}]{fang2020perceptual}
Yuming Fang, Hanwei Zhu, Yan Zeng, Kede Ma, and Zhou Wang.
\newblock Perceptual quality assessment of smartphone photography.
\newblock In {\em Proceedings of the IEEE/CVF Conference on Computer Vision and Pattern Recognition}, pages 3677--3686, 2020.

\bibitem[\protect\citeauthoryear{Ghadiyaram and Bovik}{2015}]{ghadiyaram2015massive}
Deepti Ghadiyaram and Alan~C Bovik.
\newblock Massive online crowdsourced study of subjective and objective picture quality.
\newblock {\em IEEE Transactions on Image Processing}, 25(1):372--387, 2015.

\bibitem[\protect\citeauthoryear{Golestaneh \bgroup \em et al.\egroup }{2022}]{TReS}
S~Alireza Golestaneh, Saba Dadsetan, and Kris~M Kitani.
\newblock No-reference image quality assessment via transformers, relative ranking, and self-consistency.
\newblock In {\em Proceedings of the IEEE/CVF Winter Conference on Applications of Computer Vision}, pages 1220--1230, 2022.

\bibitem[\protect\citeauthoryear{He \bgroup \em et al.\egroup }{2016}]{resnet}
Kaiming He, Xiangyu Zhang, Shaoqing Ren, and Jian Sun.
\newblock Deep residual learning for image recognition.
\newblock In {\em Proceedings of the IEEE conference on computer vision and pattern recognition}, pages 770--778, 2016.

\bibitem[\protect\citeauthoryear{Hendrycks and Dietterich}{2019}]{hendrycks2019benchmarking}
Dan Hendrycks and Thomas Dietterich.
\newblock Benchmarking neural network robustness to common corruptions and perturbations.
\newblock {\em arXiv preprint arXiv:1903.12261}, 2019.

\bibitem[\protect\citeauthoryear{Ho \bgroup \em et al.\egroup }{2020}]{ho2020denoising}
Jonathan Ho, Ajay Jain, and Pieter Abbeel.
\newblock Denoising diffusion probabilistic models.
\newblock {\em Advances in neural information processing systems}, 33:6840--6851, 2020.

\bibitem[\protect\citeauthoryear{Hosu \bgroup \em et al.\egroup }{2020}]{hosu2020koniq}
Vlad Hosu, Hanhe Lin, Tamas Sziranyi, and Dietmar Saupe.
\newblock Koniq-10k: An ecologically valid database for deep learning of blind image quality assessment.
\newblock {\em IEEE Transactions on Image Processing}, 29:4041--4056, 2020.

\bibitem[\protect\citeauthoryear{Hou \bgroup \em et al.\egroup }{2014}]{hou2014blind}
Weilong Hou, Xinbo Gao, Dacheng Tao, and Xuelong Li.
\newblock Blind image quality assessment via deep learning.
\newblock {\em IEEE transactions on neural networks and learning systems}, 26(6):1275--1286, 2014.

\bibitem[\protect\citeauthoryear{Huang \bgroup \em et al.\egroup }{2023}]{huang2023knowledge}
Tao Huang, Yuan Zhang, Mingkai Zheng, Shan You, Fei Wang, Chen Qian, and Chang Xu.
\newblock Knowledge diffusion for distillation.
\newblock {\em arXiv preprint arXiv:2305.15712}, 2023.

\bibitem[\protect\citeauthoryear{Ke \bgroup \em et al.\egroup }{2021}]{ke2021musiq}
Junjie Ke, Qifei Wang, Yilin Wang, Peyman Milanfar, and Feng Yang.
\newblock Musiq: Multi-scale image quality transformer.
\newblock In {\em Proceedings of the IEEE/CVF International Conference on Computer Vision}, pages 5148--5157, 2021.

\bibitem[\protect\citeauthoryear{Kim and Lee}{2016}]{BIECON}
Jongyoo Kim and Sanghoon Lee.
\newblock Fully deep blind image quality predictor.
\newblock {\em IEEE Journal of selected topics in signal processing}, 11(1):206--220, 2016.

\bibitem[\protect\citeauthoryear{Larson and Chandler}{2010}]{larson2010most}
Eric~Cooper Larson and Damon~Michael Chandler.
\newblock Most apparent distortion: full-reference image quality assessment and the role of strategy.
\newblock {\em Journal of electronic imaging}, 19(1):011006, 2010.

\bibitem[\protect\citeauthoryear{Li \bgroup \em et al.\egroup }{2009}]{li2009natural}
Xuelong Li, Dacheng Tao, Xinbo Gao, and Wen Lu.
\newblock A natural image quality evaluation metric.
\newblock {\em Signal Processing}, 89(4):548--555, 2009.

\bibitem[\protect\citeauthoryear{Lin \bgroup \em et al.\egroup }{2019}]{lin2019kadid}
Hanhe Lin, Vlad Hosu, and Dietmar Saupe.
\newblock Kadid-10k: A large-scale artificially distorted iqa database.
\newblock In {\em 2019 Eleventh International Conference on Quality of Multimedia Experience (QoMEX)}, pages 1--3. IEEE, 2019.

\bibitem[\protect\citeauthoryear{Liu \bgroup \em et al.\egroup }{2017}]{Liu_2017_ICCV}
Xialei Liu, Joost van~de Weijer, and Andrew~D. Bagdanov.
\newblock Rankiqa: Learning from rankings for no-reference image quality assessment.
\newblock In {\em The IEEE International Conference on Computer Vision (ICCV)}, Oct 2017.

\bibitem[\protect\citeauthoryear{Ma \bgroup \em et al.\egroup }{2017}]{MEON}
Kede Ma, Wentao Liu, Kai Zhang, Zhengfang Duanmu, Zhou Wang, and Wangmeng Zuo.
\newblock End-to-end blind image quality assessment using deep neural networks.
\newblock {\em IEEE Transactions on Image Processing}, 27(3):1202--1213, 2017.

\bibitem[\protect\citeauthoryear{Mittal \bgroup \em et al.\egroup }{2012}]{BRISQUE}
Anish Mittal, Anush~Krishna Moorthy, and Alan~Conrad Bovik.
\newblock No-reference image quality assessment in the spatial domain.
\newblock {\em IEEE Transactions on image processing}, 21(12):4695--4708, 2012.

\bibitem[\protect\citeauthoryear{Pan \bgroup \em et al.\egroup }{2022}]{pan2022dacnn}
Zhaoqing Pan, Hao Zhang, Jianjun Lei, Yuming Fang, Xiao Shao, Nam Ling, and Sam Kwong.
\newblock Dacnn: Blind image quality assessment via a distortion-aware convolutional neural network.
\newblock {\em IEEE Transactions on Circuits and Systems for Video Technology}, 32(11):7518--7531, 2022.

\bibitem[\protect\citeauthoryear{Ponomarenko \bgroup \em et al.\egroup }{2015}]{ponomarenko2015image}
Nikolay Ponomarenko, Lina Jin, Oleg Ieremeiev, Vladimir Lukin, Karen Egiazarian, Jaakko Astola, Benoit Vozel, Kacem Chehdi, Marco Carli, Federica Battisti, et~al.
\newblock Image database tid2013: Peculiarities, results and perspectives.
\newblock {\em Signal processing: Image communication}, 30:57--77, 2015.

\bibitem[\protect\citeauthoryear{Qin \bgroup \em et al.\egroup }{2023}]{DEIQT}
Guanyi Qin, Runze Hu, Yutao Liu, Xiawu Zheng, Haotian Liu, Xiu Li, and Yan Zhang.
\newblock Data-efficient image quality assessment with attention-panel decoder.
\newblock In {\em Proceedings of the Thirty-Seventh AAAI Conference on Artificial Intelligence}, 2023.

\bibitem[\protect\citeauthoryear{Radford \bgroup \em et al.\egroup }{2021}]{radford2021learning}
Alec Radford, Jong~Wook Kim, Chris Hallacy, Aditya Ramesh, Gabriel Goh, Sandhini Agarwal, Girish Sastry, Amanda Askell, Pamela Mishkin, Jack Clark, et~al.
\newblock Learning transferable visual models from natural language supervision.
\newblock In {\em International conference on machine learning}, pages 8748--8763. PMLR, 2021.

\bibitem[\protect\citeauthoryear{Rombach \bgroup \em et al.\egroup }{2022}]{rombach2022high}
Robin Rombach, Andreas Blattmann, Dominik Lorenz, Patrick Esser, and Bj{\"o}rn Ommer.
\newblock High-resolution image synthesis with latent diffusion models.
\newblock In {\em Proceedings of the IEEE/CVF conference on computer vision and pattern recognition}, pages 10684--10695, 2022.

\bibitem[\protect\citeauthoryear{Saad \bgroup \em et al.\egroup }{2012}]{saad2012blind}
Michele~A Saad, Alan~C Bovik, and Christophe Charrier.
\newblock Blind image quality assessment: A natural scene statistics approach in the dct domain.
\newblock {\em IEEE transactions on Image Processing}, 21(8):3339--3352, 2012.

\bibitem[\protect\citeauthoryear{Selvaraju \bgroup \em et al.\egroup }{2017}]{grad}
Ramprasaath~R Selvaraju, Michael Cogswell, Abhishek Das, Ramakrishna Vedantam, Devi Parikh, and Dhruv Batra.
\newblock Grad-cam: Visual explanations from deep networks via gradient-based localization.
\newblock In {\em Proceedings of the IEEE international conference on computer vision}, pages 618--626, 2017.

\bibitem[\protect\citeauthoryear{Sheikh \bgroup \em et al.\egroup }{2006}]{sheikh2006statistical}
Hamid~R Sheikh, Muhammad~F Sabir, and Alan~C Bovik.
\newblock A statistical evaluation of recent full reference image quality assessment algorithms.
\newblock {\em IEEE Transactions on image processing}, 15(11):3440--3451, 2006.

\bibitem[\protect\citeauthoryear{Shi and Lin}{2020}]{shi2020full}
Chenyang Shi and Yandan Lin.
\newblock Full reference image quality assessment based on visual salience with color appearance and gradient similarity.
\newblock {\em IEEE Access}, 8:97310--97320, 2020.

\bibitem[\protect\citeauthoryear{Song \bgroup \em et al.\egroup }{2020}]{ddim}
Jiaming Song, Chenlin Meng, and Stefano Ermon.
\newblock Denoising diffusion implicit models.
\newblock {\em arXiv preprint arXiv:2010.02502}, 2020.

\bibitem[\protect\citeauthoryear{Song \bgroup \em et al.\egroup }{2023}]{song2023active}
Tianshu Song, Leida Li, Deqiang Cheng, Pengfei Chen, and Jinjian Wu.
\newblock Active learning-based sample selection for label-efficient blind image quality assessment.
\newblock {\em IEEE Transactions on Circuits and Systems for Video Technology}, 2023.

\bibitem[\protect\citeauthoryear{Su \bgroup \em et al.\egroup }{2020}]{hypernet}
Shaolin Su, Qingsen Yan, Yu~Zhu, Cheng Zhang, Xin Ge, Jinqiu Sun, and Yanning Zhang.
\newblock Blindly assess image quality in the wild guided by a self-adaptive hyper network.
\newblock In {\em Proceedings of the IEEE/CVF Conference on Computer Vision and Pattern Recognition}, pages 3667--3676, 2020.

\bibitem[\protect\citeauthoryear{Tao \bgroup \em et al.\egroup }{2009}]{tao2009reduced}
Dacheng Tao, Xuelong Li, Wen Lu, and Xinbo Gao.
\newblock Reduced-reference iqa in contourlet domain.
\newblock {\em IEEE Transactions on Systems, Man, and Cybernetics, Part B (Cybernetics)}, 39(6):1623--1627, 2009.

\bibitem[\protect\citeauthoryear{Touvron \bgroup \em et al.\egroup }{2022}]{touvron2022deit}
Hugo Touvron, Matthieu Cord, and Herv{\'e} J{\'e}gou.
\newblock Deit iii: Revenge of the vit.
\newblock {\em arXiv preprint arXiv:2204.07118}, 2022.

\bibitem[\protect\citeauthoryear{Wang \bgroup \em et al.\egroup }{2004}]{wang2004image}
Zhou Wang, Alan~C Bovik, Hamid~R Sheikh, and Eero~P Simoncelli.
\newblock Image quality assessment: from error visibility to structural similarity.
\newblock {\em IEEE transactions on image processing}, 13(4):600--612, 2004.

\bibitem[\protect\citeauthoryear{Wu \bgroup \em et al.\egroup }{2020}]{end}
Jinjian Wu, Jupo Ma, Fuhu Liang, Weisheng Dong, Guangming Shi, and Weisi Lin.
\newblock End-to-end blind image quality prediction with cascaded deep neural network.
\newblock {\em IEEE Transactions on image processing}, 29:7414--7426, 2020.

\bibitem[\protect\citeauthoryear{Yang \bgroup \em et al.\egroup }{2020}]{yang2020ttl}
Xiaohan Yang, Fan Li, and Hantao Liu.
\newblock Ttl-iqa: Transitive transfer learning based no-reference image quality assessment.
\newblock {\em IEEE Transactions on Multimedia}, 23:4326--4340, 2020.

\bibitem[\protect\citeauthoryear{Yang \bgroup \em et al.\egroup }{2022}]{MGD}
Zhendong Yang, Zhe Li, Mingqi Shao, Dachuan Shi, Zehuan Yuan, and Chun Yuan.
\newblock Masked generative distillation.
\newblock In {\em European Conference on Computer Vision}, pages 53--69. Springer, 2022.

\bibitem[\protect\citeauthoryear{Ying \bgroup \em et al.\egroup }{2020}]{ying2020patches}
Zhenqiang Ying, Haoran Niu, Praful Gupta, Dhruv Mahajan, Deepti Ghadiyaram, and Alan Bovik.
\newblock From patches to pictures (paq-2-piq): Mapping the perceptual space of picture quality.
\newblock In {\em Proceedings of the IEEE/CVF Conference on Computer Vision and Pattern Recognition}, pages 3575--3585, 2020.

\bibitem[\protect\citeauthoryear{You and Korhonen}{2021}]{TIQA}
Junyong You and Jari Korhonen.
\newblock Transformer for image quality assessment.
\newblock In {\em 2021 IEEE International Conference on Image Processing (ICIP)}, pages 1389--1393. IEEE, 2021.

\bibitem[\protect\citeauthoryear{Zhang \bgroup \em et al.\egroup }{2015}]{ILNIQE}
Lin Zhang, Lei Zhang, and Alan~C Bovik.
\newblock A feature-enriched completely blind image quality evaluator.
\newblock {\em IEEE Transactions on Image Processing}, 24(8):2579--2591, 2015.

\bibitem[\protect\citeauthoryear{Zhang \bgroup \em et al.\egroup }{2018}]{zhang2018blind}
Weixia Zhang, Kede Ma, Jia Yan, Dexiang Deng, and Zhou Wang.
\newblock Blind image quality assessment using a deep bilinear convolutional neural network.
\newblock {\em IEEE Transactions on Circuits and Systems for Video Technology}, 30(1):36--47, 2018.

\bibitem[\protect\citeauthoryear{Zhang \bgroup \em et al.\egroup }{2022}]{zhang2022contrastive}
Linfeng Zhang, Xin Chen, Junbo Zhang, Runpei Dong, and Kaisheng Ma.
\newblock Contrastive deep supervision.
\newblock In {\em Computer Vision--ECCV 2022: 17th European Conference, Tel Aviv, Israel, October 23--27, 2022, Proceedings, Part XXVI}, pages 1--19. Springer, 2022.

\bibitem[\protect\citeauthoryear{Zhang \bgroup \em et al.\egroup }{2023}]{zhang2023blind}
Weixia Zhang, Guangtao Zhai, Ying Wei, Xiaokang Yang, and Kede Ma.
\newblock Blind image quality assessment via vision-language correspondence: A multitask learning perspective.
\newblock In {\em Proceedings of the IEEE/CVF Conference on Computer Vision and Pattern Recognition}, pages 14071--14081, 2023.

\bibitem[\protect\citeauthoryear{Zhao \bgroup \em et al.\egroup }{2023}]{QPT}
Kai Zhao, Kun Yuan, Ming Sun, Mading Li, and Xing Wen.
\newblock Quality-aware pre-trained models for blind image quality assessment.
\newblock In {\em Proceedings of the IEEE/CVF Conference on Computer Vision and Pattern Recognition}, pages 22302--22313, 2023.

\bibitem[\protect\citeauthoryear{Zhou \bgroup \em et al.\egroup }{2019}]{no}
Yu~Zhou, Leida Li, Shiqi Wang, Jinjian Wu, Yuming Fang, and Xinbo Gao.
\newblock No-reference quality assessment for view synthesis using dog-based edge statistics and texture naturalness.
\newblock {\em IEEE Transactions on Image Processing}, 28(9):4566--4579, 2019.

\bibitem[\protect\citeauthoryear{Zhu \bgroup \em et al.\egroup }{2020}]{zhu2020metaiqa}
Hancheng Zhu, Leida Li, Jinjian Wu, Weisheng Dong, and Guangming Shi.
\newblock Metaiqa: Deep meta-learning for no-reference image quality assessment.
\newblock In {\em Proceedings of the IEEE/CVF Conference on Computer Vision and Pattern Recognition}, pages 14143--14152, 2020.

\end{thebibliography}

\end{document}